\documentclass[sigconf]{acmart} 

\usepackage[utf8]{inputenc} 
\usepackage[T1]{fontenc}    
\usepackage{hyperref}       
\usepackage{amsfonts}       
\usepackage{amsmath}

\usepackage{algorithm}
\usepackage{algorithmic}

\AtBeginDocument{%
  \providecommand\BibTeX{{%
    \normalfont B\kern-0.5em{\scshape i\kern-0.25em b}\kern-0.8em\TeX}}}

\renewcommand\footnotetextcopyrightpermission[1]{}

\usepackage{etoolbox}
\makeatletter
\patchcmd{\maketitle}{\@copyrightspace}{}{}{}
\makeatother

\begin{document}

\title{Learn to Bind and Grow Neural Structures}

\author{Azhar Shaikh}
\affiliation{\institution{PES University}}
\email{azhar199865@gmail.com}

\author{Nishant Sinha}
\affiliation{\institution{OffNote Labs}}
\email{nishant@offnote.co}

\renewcommand{\shortauthors}{Shaikh and Sinha}

\begin{abstract}
  Task-incremental learning involves the challenging problem of learning new tasks continually, without forgetting past knowledge. Many approaches address the problem by expanding the structure of a shared neural network as tasks arrive, but struggle to grow optimally, without losing past knowledge. We present a new framework, Learn to Bind and Grow, which learns a neural architecture for a new task incrementally, either by binding with layers of a similar task or by expanding layers which are more likely to conflict between tasks. Central to our approach is a novel, interpretable, parameterization of the shared, multi-task architecture space, which then enables computing globally optimal architectures using Bayesian optimization. Experiments on continual learning benchmarks show that our framework performs comparably with earlier expansion based approaches and is able to flexibly compute multiple optimal solutions with performance-size trade-offs.
\end{abstract}



    \label{fig:overview}

\maketitle

\pagestyle{plain}


\newcommand{\tasks}{T}
\newcommand{\tseq}{[\tasks]}
\newcommand{\trainds}{\mathcal{D}}
\newcommand{\valds}{\mathcal{V}}
\newcommand{\testds}{\mathcal{T}}
\newcommand{\loss}{\mathcal{L}}
\newcommand{\params}{\Theta}
\newcommand{\param}{\theta}
\newcommand{\modelf}{f}

\newcommand{\layers}{L}
\newcommand{\snet}{\mathcal{G}} 
\newcommand{\inet}{{\snet}^{\downarrow}}
\newcommand{\lcf}{{\mathcal C}} 
\newcommand{\cfp}{\tilde{\lcf}} 
\newcommand{\cf}{{\mathbb C}} 
\newcommand{\grt}{\delta} 

\newcommand{\gseq}{\pi}
\newcommand{\Gseq}{\Pi} 
\newcommand{\gseqvec}{\hat{\tau}}
\newcommand{\valerr}{\mathcal{E}}

\newcommand{\pcoeff}{\lambda}
\newcommand{\acq}{I}
\newcommand{\unif}{{\mathcal U}}
\newcommand{\normal}{N} 
\newcommand{\cumnormal}{\Psi}
\newcommand{\real}{{\mathcal R}}

\newcommand{\lnorm}[1]{\left\lVert#1\right\rVert}


\section{Introduction}

A key challenge for efficient learning systems is to learn new information without forgetting information gained previously.
Solving this challenge is crucial for deep neural networks (DNNs), which seek to continue learning new tasks without {\it catastrophically} forgetting old ones~\cite{mcohen89,ratcliff90,robins95,thrunLL}.
Because the participating tasks are varied and may have conflicting objectives, we can view task-incremental learning as a complex, multi-objective optimization problem~\cite{multi-objective-koltun}.
Broadly, two main approaches have been proposed to address this problem.
The first set of methods impose some form of regularization on a fixed capacity network, by restricting changes to important parameters of each task~\cite{ewc,zenke-synaptic,imm} or using constrained experience replay~\cite{gem,mer}.
The first set of methods impose some form of regularization on a fixed capacity network, by restricting changes to important parameters of each task~\cite{ewc,zenke-synaptic,imm}, using constrained experience replay~\cite{gem,mer} or distillation~\cite{pandc}.
The second set of methods tackle forgetting by dynamic network expansion, i.e., adding additional task-specific parameters when a new task arrives.
The challenge here is to minimize the expansion and avoid exploding the network parameters during learning.
In other words, the expansion based approaches deal with the {\it expand-vs-forget} dilemma, a variant of the classical stability-plasticity dilemma~\cite{stability-plasticity}.

Existing approaches~\cite{pgn,den,rcl,l2g,apd20} address the dilemma in different ways.
DEN~\cite{den} adds new fine-grained neurons selectively per layer, and minimizes expansion by using group sparsity regularization on newly added parameters.
PGN~\cite{pgn} {\it freezes} the learned parameters of the earlier tasks and uses lateral connections to reuse old parameters for the new task.
RCL~\cite{rcl} uses reinforcement learning to compute the number of channels that should be added to each network layer, when a new task arrives for training.
APD~\cite{apd20} constrains network parameters to be a sum of shared and (sparse-) task specific ones, and further decomposes and clusters task specific parameters to minimize expansion.
The Learn-to-grow~\cite{l2g} method, which is closely related, specifies that each network layer may either be reused, adapted or cloned for the new task, and performs differentiable neural architecture search (DNAS)~\cite{DARTS} over a combined architecture to ascertain the optimal expansion choice for each layer.
Although all existing methods trade-off between growth and interference in different ways, none of them directly exploit {\it task similarity} or {\it layer similarity} for selective expansion.
They either expand upon the previous network as a whole or add fine-grained neuron units.
In contrast, we propose to {\it identify} an existing coarser-level task {\it sub-network}, which already computes features relevant to the new task, and reuse it for incremental learning.
Moreover, existing methods perform per-task optimization only, without considering global optimization across the task set.
Global optimization enables finding networks with better performance-size tradeoffs, which task-local optimization can fail to find. 

We present {\it Learn to Bind and Grow} (L2BG), a new expansion based, task-incremental learning approach, which addresses the above issues.
L2BG exploits the observation that in most task groups learned together, the tasks are mutually similar to different degrees. Hence, to learn a new task $t$, we find a previously learned task $b$ which is similar to $t$ ({\it bind}), and then build the task network for $t$ upon the learned network structure for $b$ ({\it grow}), without modifying the rest of the network structure.
By considering only a small set of layers from $b$ for expansion, we curtail network growth as well as boost learning efficiency, while retaining the task-specific performance.
\begin{figure*}[h]
    \centering
    \includegraphics[width=14cm]{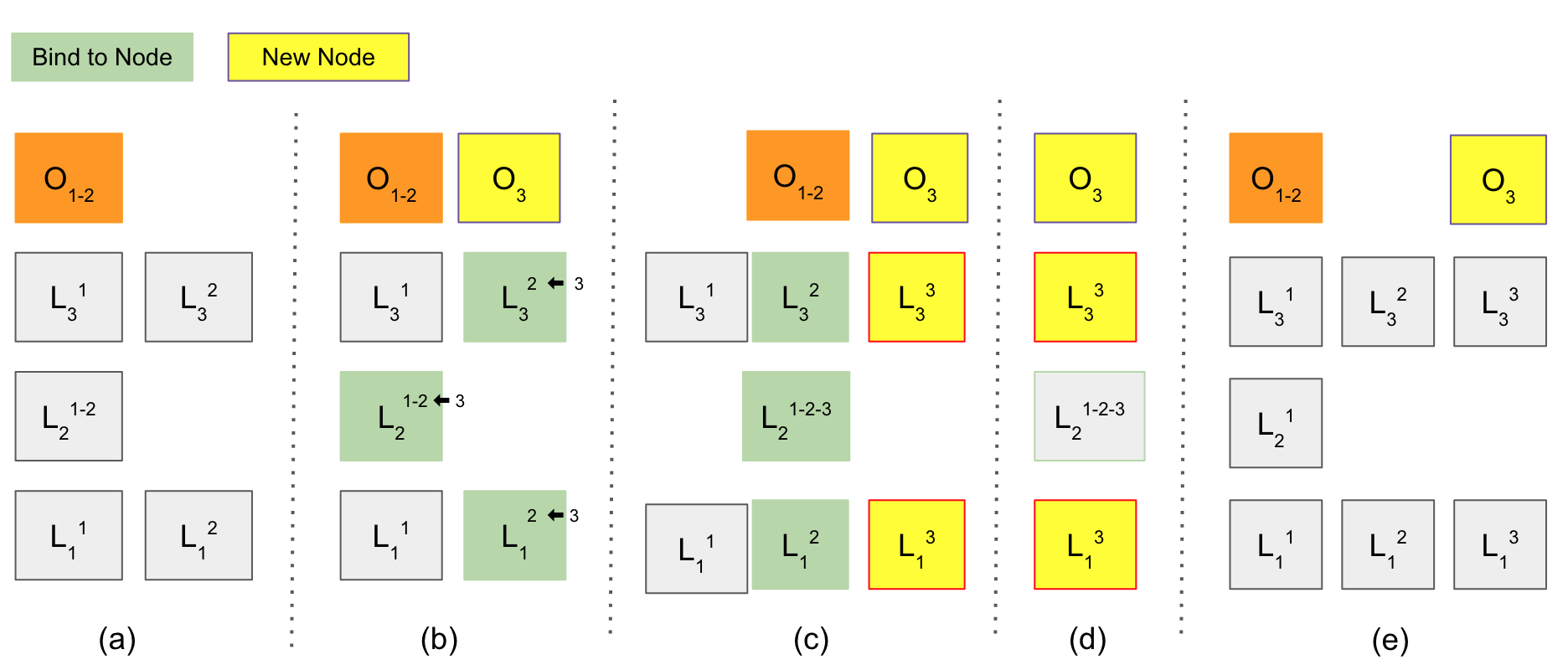}
    \caption{Illustration of the proposed learn-to-bind-grow framework. $L^t_l$: $l$ denotes the layer index and $t$ the task indices sharing this layer instance. (a) Initial Joint Network: task 1 and 2 each have their own instances of layer 1 and 3, while layer 2 is shared (b) New task 3 picks task 2 (and its layers) to bind with. (c) The conflict model is estimated and conflicting layers (1 and 3) expanded, $L_2$ remains shared (d) The new task-net for task 3, including new and shared layers, is trained. e) The updated Joint Network after task 3 is trained
    and its new layers incorporated.}
    \label{fig:overview}
\end{figure*}
L2BG maintains a joint, parameterized network $\snet$ for learning multiple tasks, with both task-specific and shared layers.
For processing a new task $t$, we first identify an existing trained task $b$ in $\snet$, which is similar to $t$ and then bind $t$ to $b$.
Binding implies that tasks $t$ and $b$ now share the same sub-network $\snet_b$ of $\snet$.
Next, we identify one or more layers $l$ of $\snet_b$, on which $t$ and $b$ may {\it conflict}, i.e., sharing $l$ between $t$ and $b$ may cause catastrophic forgetting for the tasks.
We then remove the conflict by {\it growing} a new corresponding layer $l'$ for $\snet_t$, and re-train $\snet_t$ for $t$.
Both task binding and layer expansion rely on a novel, hierarchical, {\it conflict estimation} model proposed in this paper. 
The model makes the expansion process {\it interpretable}: we can explain the task bindings and layer expansions choices precisely, using the conflict model scores.
Fig.~\ref{fig:overview} illustrates the proposed algorithm.

We show that our bind-and-grow strategy induces a new parameterization of the space of shared, multi-task architectures, which is linear in the number of tasks.
This, in turn, allows us to employ Bayesian Optimization~\cite{raswi} to search for optimal network architectures, under multiple objectives: maximize the average task performance and keep the joint network size low.
Our parameterization enables exploring the space of performance-size tradeoffs by constraining global optimization {\it flexibly}, to be effective even in low budget scenarios.

In summary, our contributions are:
\begin{itemize}
\itemsep0em 
\item A new method to perform task-incremental learning, which reuses previously learned information efficiently by binding to specific trained sub-networks, and dynamically expands conflicting layers to avoid catastrophic forgetting among multiple tasks.
\item Unlike existing expansion based methods, our method exploits task similarity and includes an interpretable criteria for dynamic expansions based on an hierarchical conflict model.
\item A new parameterization of the multi-task architecture space, which enables efficient multi-objective Bayesian Optimization to find Pareto optimal solutions~\cite{roijers13, multi-objective-koltun}.
\end{itemize}

\section{Preliminaries}
\label{sec:prelims}

Consider a large collection of tasks $\tasks$ to be learned.
Each task $t\in\tasks$ has train, validation and test datasets, $\trainds_t$, $\valds_t$, $\testds_t$, respectively.
Individually, each task has a learnable model $\modelf(\cdot;\param_t)$, parameterized over $\param_t$, a loss function $\loss_t$. 
The model is represented as a neural network graph, with parameters distributed among layers, which are drawn from a fixed set $\layers$.

In this paper, we consider two scenarios: (a) training data for task $t$ is not available after $t$ is trained and (b) training data for all tasks is available throughout, but training is incremental per task.
In the first scenario, L2BG performs task-incremental optimization only, whereas in the second one, it performs global optimization across tasks also.
Several large-scale multi-task problems across text, vision and RL domains~\cite{superglue,mmnmt,taskonomy,metaworld} fall into the latter scenario.

{\bf JointNet.} During task-incremental learning, we construct a joint neural network graph for tasks in $\tasks$, called {\it JointNet} $\snet$, which contains task-specific as well as shared layers between two or more tasks.
Formally, the JointNet is a directed acyclic graph (DAG) $\snet=(N, E)$, where nodes $N$ correspond to a layer $l \in \layers$, and edges $E$ connect the layers. 
The DAG $\snet$ has a single source node (input) and multiple sink nodes (outputs), where each sink node corresponds to one or more tasks.
For each tasks $t\in\tasks$, we can select a sub-graph $\snet_t$ from $\snet$; $\snet_t$ may share nodes with sub-graphs $\snet_{t'}$ of other tasks $t'$. We call $\snet_t$ as the {\it task-net} for $t$ and that $\snet$ {\it contains} a task-net for each $t$.
In this paper, we assume that task-nets for all tasks $t\in\tasks$ are similar: there exists a {\it bijective} mapping between layers of each pair of tasks.
We use the train dataset $\trainds_t$ for training each task only once; the validation datasets $\valds_t$ are used for evaluation throughout the learning.
Fig.~\ref{fig:overview}(a) and (e) depict the JointNet layers before and after task 3 arrives, respectively.

We formalize the case when each task $t$ is trained independently, using a {\it degenerate} JointNet $\inet$. 
Here, the task sub-graph $\inet_t$ for $t$ does not share any nodes with sub-graph $\inet_{t'}$ for all $t'\neq t$. The network $\inet_t$, called the {\it independent} task-net for $t$, is used to train task $t$ in isolation. For a new task $t$, we make use of $\inet_t$ to discover which task $t'$ to bind $t$ with.

{\bf Multi-task Optimization.} The global multi-task optimization problem involves minimizing the following loss $\loss$:
$$ \loss = \sum_{t\in\tasks} \loss_t(\param_t; \trainds_t) $$
where $\loss_t$ is the individual loss for task $t$ computed over the training set $\trainds_t$.
Optimizing the global loss $\loss$ directly may be infeasible when $\tasks$ is large or if all tasks are not available simultaneously.
In such cases, we resort to {\it task-incremental} optimization, i.e., optimize $\loss_t(\params_t;\cdot)$ when task $t$ arrives.
Because tasks may share parameters, optimizing for task $t$ may affect the performance of the set of tasks $P_t$ arrived before $t$. The goal of our work is grow the existing JointNet to avoid catastrophic forgetting so that the combined loss ($\sum_{t'\in P_t} \loss_{t'}$) for existing tasks does not change significantly.

\section{Learn to Bind and Grow}
\label{sec:l2bg}

We now describe the proposed method, Learn to Bind and Grow.
The method maintains a shared, multi-task, model JointNet $\snet$, for a set of tasks $\tasks$, containing a task-net for each $t'\in\tasks$,
When a new task $t$ arrives, the method {\it binds} the task-net for $t$ to the task-net of an existing, {\it similar} task $b$.
Intuitively, this means that the task-net layers of $b$ and $t$ are shared initially, ignoring any difference in the task-specific architectures of the two tasks.
In the {\it grow} step, the method expands the task-net for $b$ by adding new task-specific layers for $t$, and re-routes the task-net of $t$ to use the new layers.
There are two key problems here: (a) which task $b$ to bind $t$ with, (b) which layers of $b$ to grow.

\noindent{\bf Hierarchical Conflict Model.} We propose a novel, hierarchical, {\it conflict model} to address both problems.
The model computes layer-wise conflict scores $\lcf$ between the task-nets of $t$ and each task $b$ from the set of bind {\it candidates} $B$.
These scores play a dual role.
First, by aggregating pairwise layer scores to pairwise task scores, we determine which task $b\in B$ conflicts the {\it least} with $t$; $t$ binds to a task $b$ with the least conflict score.
Second, given a layer $l$ in the task net of $b$, the scores indicate the {\it likelihood} of $b$ and $t$ to {\it interfere} on $l$. The higher the score, more likely the interference and so $l$ cannot be shared between $b$ and $t$.

In this paper, we assume that the structure of task-nets across tasks is similar, i.e., we have a bijective mapping between layers of individual task-nets for each task pair ($t$, $b$).
Let $l$ index over the layer pairs $L$ in the mapping.
We denote the layer-wise conflict score for a layer $l\in L$ as $\lcf_{t,b}(l)\in (0,1)$.

\noindent{\bf Binding.} Given a set of candidate tasks $b'\in B$ and the layer-wise conflict scores $\lcf_{t,b'}(l)$, represented as vector ${\bf\lcf_{t,b'}}$ we compute the pair-wise task conflict scores $\cf_{t, b'}$ between $t$ and $b'$ as
$ \cf_{t, b'} = \lnorm{\bf \lcf_{t,b'}} $.
Now, task $t$ binds to a candidate task $b\in B$ with the least $\cf_{t, b}$ score.
Binding implies that the task-nets for $t$ and $b$ are shared; hence we also say that $\snet_t$ binds with $\snet_b$.

{\bf Grow-Step.} Once we bind the task-net $\snet_t$ for $t$ to the task-net $\snet_b$ of a prior task $b$, we again use $\lcf_{t,b}$ to determine which shared layers are more likely to cause interference, and try to {\it unshare} those layers by expanding them. 
To expand a layer $l$, we create a new copy $l_t$ of $l$, modify the incoming and outgoing edges to $l_t$ and update the task-net $\snet_t$ for task $t$, while keeping the rest of the JointNet intact.
To compute the layers for expansion, we first transform per layer conflict scores into a probability distribution $\cfp_{t,b}(l)$ over layers $l\in L$.
Now, we may {\it sample} layers from $\cfp_{t,b}$ based on their probability mass, and expand them individually.
This, however, leads to a combinatorial explosion in the space of explored architectures.

\noindent{\bf  Grow Coefficients.} To mitigate this, we introduce a new parameterization based on a {\it grow coefficient} per task.
A grow coefficient $\grt_t\in [0,1]$ for task $t$ provides an {\it dynamic} upper bound on the number of new layers created for $t$, based on the cumulative probability distribution of $\cfp$.

Given a coefficient $\grt_t$ for task $t$, we first compute a {\it nucleus}~\cite{nucleus-text} of $\cfp$ under $\grt_t$: the smallest possible set of layers, whose cumulative probability under $\cfp$ exceeds the bound $\grt_t$.
We then expand all the layers in the nucleus to obtain a new task-net $\snet_t$, which is re-trained using the training dataset $\trainds_t$ for t.
Training $\snet_t$ inside the joint network $\snet$ enables forward information transfer from bound task $b$ (and its transitive bindings), and often improves the task performance on $t$.
Note that with a single coefficient $\grt_t$ per task $t$, we can span the complete expansion space for $t$: from no expansion, full sharing ($\grt_t \sim 0$) to expanding all the layers ($\grt_t \sim 1$).
This capability comes with a caveat that all expansions must abide with the layer ordering given by the conflict distribution $\cfp$.
We summarize our {Bind-Grow-Step} algorithm in Algorithm~\ref{alg:bgstep}.

\begin{algorithm}[t]
\caption{Bind-Grow-Step Algorithm}
\label{alg:bgstep}
\small{
\begin{algorithmic}
\STATE \textbf{Input:} JointNet $\snet$, task $t$, bind candidate tasks $B$, grow coefficient $\grt_t$
\STATE \textbf{Output:} Expanded JointNet $\snet'$, Validation Errors $\valerr$

\FOR{each pair $(t,b')$, $b'\in B$}
\STATE Compute task conflict scores $\cf_{t,b'}$
\STATE Compute the layer-wise conflict distribution $\cfp_{t,b'}$ 
\ENDFOR
\STATE Let bind task $b$ := argmin$_{b'\in B}$ $\cf_{t,b'}$.
\STATE Compute the nucleus layers $L'$ using the cumulative $\cfp_{t,b}$ distribution.
\STATE Expand layers $L'$ in $\snet$ to obtain $\snet'$.
\STATE Train $\snet'_t$ using train data $\trainds_t$.
\STATE Compute validation errors $\valerr_t$ for all existing tasks $t$ using dataset $\valds_t$.
\end{algorithmic}
}
\end{algorithm}

{\bf Computing the Conflict Model.} Computing a precise, layer-wise conflict model between tasks involves characterizing interference precisely~\cite{mcohen89,ratcliff90,mer}, which is intractable~\cite{stability-plasticity}.
Several approaches to approximate interference are possible but expensive, e.g., we may identify the overlap of important parameters for both tasks using the Fisher Matrix~\cite{ewc} or try different combinations of layer expansions with re-training and evaluation.
Instead, we compute a low-cost, approximate conflict model, by correlating the layer-wise features of an independently trained task-net $\inet_t$ for $t$ (see Sec.~\ref{sec:prelims}) and the current task-net $\snet_b$ for binding candidate $b$.
We use the current (fully-trained) task-net $\snet_b$, instead of say, a pre-trained $\inet_b$, to ensure that the conflict model is computed in context of the growing network.
When a task $t$ arrives, we first train an individual network $\inet_t$ for $t$.
Then, we sample data from validation set $\valds_t$, feed to both $\snet_b$ and $\inet_t$ and collect post-activations after each layer of the networks.
Finally, using the representational similarity analysis (RSA) algorithm~\cite{RSA,rsataxonomy}, we compute correlation scores between pairwise post-activations of mapped layers $l\in L$, for every sample in $\valds_t$.
The conflict score $\lcf_{t,b}(l)$ = $\rho_l$ where $\rho_l\in[-1,1]$ denotes the RSA dissimilarity score for $l$.
Our implementation follows the original RSA algorithm~\cite{RSA} and requires space quadratic in the size of layer post-activations. 
Further, we normalize $\lcf$ to be in the range $(0,1)$.


In the above discussion, we use a {\it layer} as the basic unit of expansion.
More generally, we can partition a layer into sub-components, e.g., a group of channels, and grow one or more groups in each grow step. 
The notion of a conflict distribution extends and applies to any well-defined, hierarchical, partition of the network structure.

{\bf Grow Sequences.} By performing a bind-grow step iteratively, for each task in an incoming sequence over tasks $\tasks$, we obtain a grow sequence $\gseq = [(t, b_t, \grt_t)]_{t\in\tasks}$.
The sequence characterizes the choices made (bind task $t$ to $b_t$ and expand with grow coefficient $\grt_t$) during a particular task-incremental learning trial.
By exploring different grow sequences, we can explore different task orders, task bindings and degrees of network expansion, and optimize over the global architecture space.

{\bf Evaluation.} When a grow sequence finishes executing with output $\snet$, we compute two measures over $\snet$: the average validation error (or accuracy) over tasks and the {\it average multi-task gain} $\Delta$, based on the average per-task drop measure in~\cite{astmt19}. 
For a task $t$, let the validation error after executing a grow sequence $\gseq$ be $\valerr_{\gseq,t}$ and for the single-task baseline be $\valerr_{\phi,t}$.
The average multi-task gain $\Delta_\gseq$ is defined as
\[
\Delta_\gseq = \sum_{t\in\tasks} (\valerr_{\gseq,t} - \valerr_{\phi,i}) / \valerr_{\phi,t}
\]
 
{\bf Comparison with Expansion Approaches.} 
Compared to Learn-to-Grow~\cite{l2g}, we perform expansion relative to a similar task net, not the full network. We perform global, as opposed to per task, optimization over model structures.
We do not consider {\it adapting} a layer here; our method can be extended by allowing both the task-net $\inet_t$ and its adapted version to bind with an existing task-net. 
Compared to RCL~\cite{rcl}, our expansion relies on an approximate conflict model, which allows one or more layers (or layer components) to be expanded simultaneously, as opposed to learning how many nodes to expand per layer.
APD~\cite{apd20} assumes a shared and task-specific decomposition for each layer; our method works with the unchanged network structures directly.
Finally, we exploit task similarity scores to guide expansion; neither PGN~\cite{pgn} or DEN~\cite{den} do so.

{\bf Global Optimization.} Grow sequences based search optimization is attractive for several reasons.
First, they enable global optimization over the {\it complete} space of shared, multi-task architectures.
Next, the optimization is {\it interpretable}, i.e., we know precisely why a task binds with an older task, as well as, why a particular set of layers are expanded.
Further, the optimization is governed by a small (linear in task set size) set of hyper-parameters, namely, the binding task index, and a grow coefficient for each task.
Also, we can perform {\it constrained} optimization under low computational budgets.
For example, we may constrain search to a fixed, prior set of bind targets, or alternatively, only explore a small, curated set of grow coefficients, while computing bindings dynamically.

\newcommand{\bof}{\mathcal{F}}
{\bf Bayesian Optimization.} Exploring all grow sequences to find optimal architectures is intractable because evaluating each grow sequence is expensive.
Bayesian Optimization (BO) is a technique to optimize expensive black-box functions efficiently, by using a surrogate probabilistic model, e.g., Gaussian Processes, to model the uncertainty of the unknown function. 
For details on BO, we refer to Rasmussen and Williams~\cite{raswi}.
Additionally, our problem involves {\it multiple, competing objectives}: average multi-task gain and network parameter count, which we want to maximize and minimize, respectively.
Hence, we seek to recover the Pareto optimal solutions~\cite{roijers13, multi-objective-koltun}: solutions which cannot be improved in one objective without degrading the other objective.

In a grow sequence $\gseq$ = $[(t, b_t, \grt_t)]_{t\in \tasks}$, task identifiers $t$ and $b_t$ are (bounded) categorical choices while the coefficient $\grt_t \in [0, 1]$ is sampled from a continuous range.
Using BO, we optimize the black-box function $\bof : \Gseq \rightarrow (\real\times\real)$, which maps a growth sequence $\gseq\in\Gseq$ to a pair of outputs, the average multi-task gain and the parameter count.

\section{Experiments}
\renewcommand{\comment}[1]{}
In this section, we test the effectiveness of the proposed L2BG framework and compare with existing dynamic expansion approaches. We seek answers to the following questions.
\begin{itemize}
    \item How effective is L2BG in enabling transfer between tasks and taming catastrophic forgetting?
    \item Is binding based on task similarity better than {\it random} binding, as in earlier approaches?
    \item How effective is RSA~\cite{RSA} in computing conflict scores?
    \item Does L2BG improve over existing expansion based approaches?
    \item Do {\it grow sequences} enable exploring trade-offs between size and performance of learned networks?
    \item Does Bayesian Optimization over grow sequences yield useful architectures, even on a low budget?
\end{itemize}

{\bf Benchmarks.} We evaluate our approach on three benchmark datasets: Permuted Mnist, Split-CIFAR100 and the visual decathalon dataset(VDD). The Permuted Mnist dataset is derived from the MNIST dataset~\cite{10027939599}: each task is a simple classification problem with a unique fixed random permutation generated to shuffle the pixels of each image, while the annotated label is kept fixed. 
The Split-CIFAR100 dataset is derived from the CIFAR100 dataset with the classes divided into T tasks with each task having 100/T classes. 
The visual decathlon dataset(VDD) consists of 10 image classification tasks – ImageNet, CIFAR100, Air-craft, DPed, Textures, GTSRB, Omniglot, SVHN, UCF101 and VGG-Flowers..

For the Permuted Mnist experiments, we use a 4-layer fully-connected network with 3 feed-forward layers and the 4th layer is the shared softmax classification layer across all tasks. For split-Cifar100 we use the network used in \cite{l2g}. The network consists of 3 convolutional layers with max-pooling and ReLU activations, followed by two fully connected layers, and finally separate task heads for each incoming task. For the VDD experiments we use AlexNet pretrained on the Imagenet dataset.
We do not present results of training on multiple tasks {\it without} expansion: previous works~\cite{l2g,rcl,apd20} have shown that expansion based approaches always outperform multi-task training of a single network.
We report the accuracy/gain for each task as an average of accuracy/gain for five different trials.
\begin{figure*}[h]
    \centering
    \includegraphics[width=15cm]{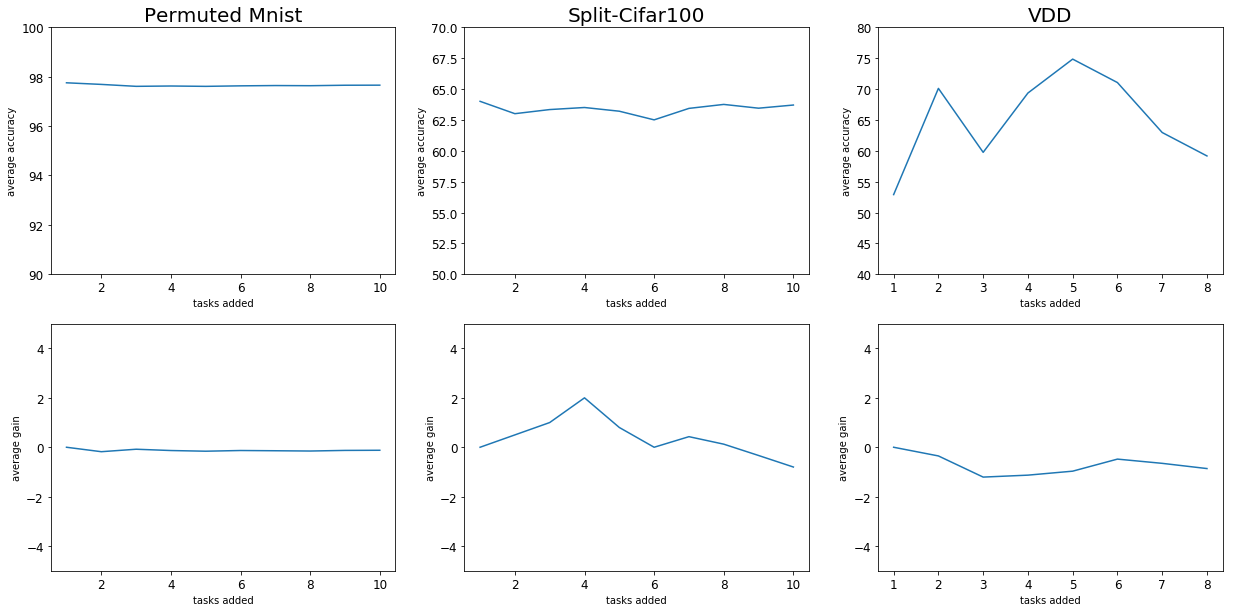}
    \caption{Average validation accuracy and Average gain plots for all datasets}
    \label{fig:avg-acc-gain}
\end{figure*}

\subsection{Evaluating Transfer and Forgetting.} 
Fig.~\ref{fig:avg-acc-gain} shows the changes in average accuracy and gain for each dataset, as the tasks learn incrementally.
Average accuracies vary depending on the performance of individual tasks: for Permuted Mnist and Split-CIFAR100, they are nearly constant, whereas they vary significantly for VDD.
This is because VDD contains multiple dissimilar tasks, e.g., aircraft vs DPed, whose final performance vary.
Average gains across all tasks has small negative values, indicating the network discovered by L2BG using dynamic expansion loses little shared information due to interference between tasks.

\begin{figure}[h]
    \centering
    \includegraphics[width=10cm]{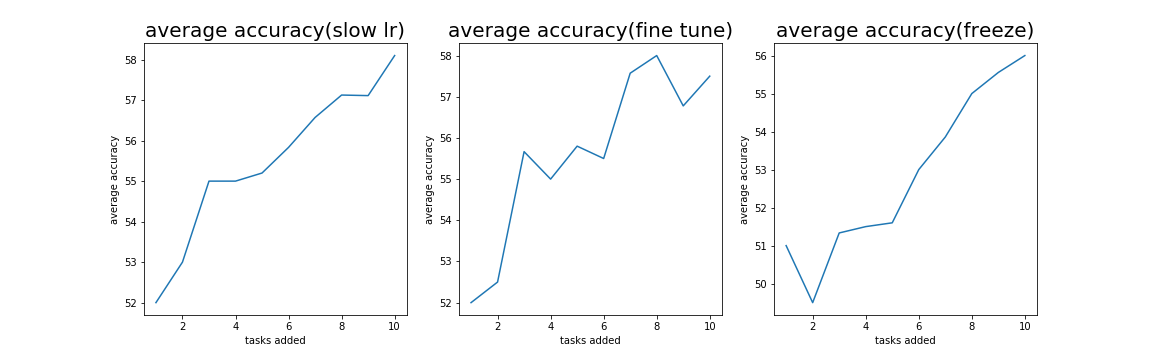}
    \caption{Different Weight Retention Strategies by varying Learning Rate (lr). freeze: lr=0, fine-tune: lr=1e-2, slow-lr: lr=1e-3. lr=1e-2.}
    \label{fig:lr}
\end{figure}

We further experiment with three {\it weight retention} options to mitigate forgetting for the {\it shared} layers: (a) freeze existing task weights (b) use reduced learning rate (lr) for existing tasks (c) no change in lr.
Freezing task weights makes it harder for the weights to adapt to the new tasks, leading to sub-optimal learning for the new tasks. In contrast, fine-tuning the shared weights allows weights to adapt to the new task but causes negative backward transfer, i.e., may increase the error on previous tasks.
As a trade-off, we allow limited backward transfer by training previous task weights with a lower learning rate.
Fig.~\ref{fig:lr} compares the three strategies on the Split-CIFAR100 dataset.
We observe that the {\it slow-lr} configuration leads to the highest final accuracy and achieves the most stable training across all configurations.

To summarize, we note that while layer expansion is effective in mitigating forgetting at the {\it coarse}-level, tweaking the learning rate for existing tasks minimizes interference further and may lead to further gains.
We plan to investigate strategies for selecting reduced learning rate and combining L2BG with fine-grained regularization methods~\cite{ewc,gem} in the future.



\begin{figure}[hbt]
    \centering
    \includegraphics[width=9cm]{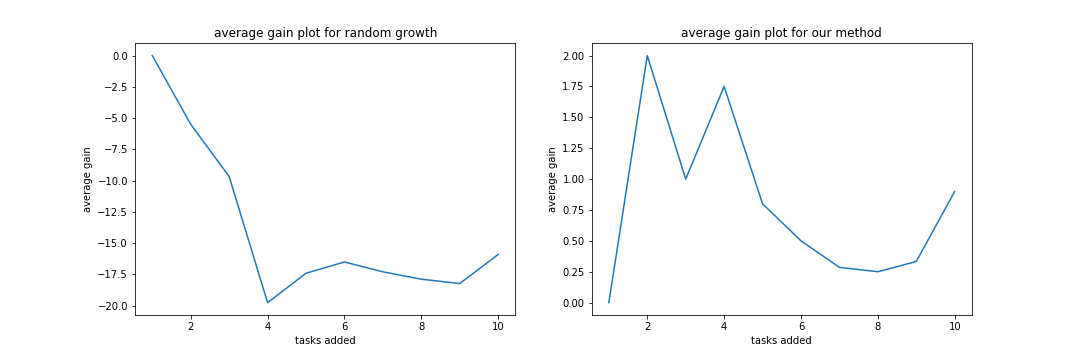}
    \caption{Average gain graph for Random growth experiment.}
    \label{fig:randomG}
\end{figure}

{\bf Is RSA effective for conflict modeling?}~Using Representation Similarity Analysis~\cite{RSA} (see Sec.~\ref{sec:l2bg}) to model layer-wise conflict distribution, helps us in two ways: first, we obtain an ordering on the conflict {\it likelihood} of the layers, and second, we find the preferred task to bind with.
Without the conflict distribution, the space of possible bindings and layer expansions is too large.
To investigate the difficulty of finding an optimal shared network, we perform a {\it random growth} experiment with the Split-Cifar100 dataset as follows.
When a new task arrives, we bind it with a {\it random} existing task, {\it randomly} choose the number of layers to expand and a subset of layers to expand.
As shown in Fig~\ref{fig:randomG}, the network obtained through random growth performs much worse when compared to the network optimized using RSA.
In fact, the average gain values go down to -20 with random-growth as compared to {\it positive} values obtained with RSA.
We therefore conclude that RSA-based conflict distributions are crucial for efficient task binding and layer expansion in L2BG.

\subsection{Final Network Representations}
We describe the final network structures obtained after training for each dataset.
We represent each Joint Network (containing task-nets for individual tasks) in a {\it tabular} format, where the columns denote the individual layer names and the {\it sequence} of layers, and each row corresponds to a particular task.
For example, the row

\verb task_4:  *, *, 3, 3, 3, \verb 

denotes that in the task-net for {\bf task 4}, the first two layers are {\bf new} (represented as "*"), while the rest of the layers {\bf bind} to {\bf task 3}.

{\bf  Split-CIFAR100 dataset.}

Here, most tasks create individual first two layers, while only some create a new third ({\it conv3}) or fourth ({\it linear1}) layer.
Most tasks share the top 2 layers ({\it linear1} and {\it linear2}).
Note that although all tasks share the top-most layer {\it linear2}, the task identifiers corresponding to {\it linear2} are different across rows.
This is because of {\it transitive} binding: {\bf task 4} binds to {\bf task 3}'s {\it linear2} layer, which in turn, binds to {\bf task 1}'s {\it linear2} layer.
\begin{verbatim}
columns: [conv1 conv2 conv3 linear1 linear2]
    task_1:  *, *, *, *, *,
    task_2:  *, *, 1, 1, 1, 
    task_3:  *, *, *, 1, 1, 
    task_4:  *, *, 3, 3, 3, 
    task_5:  *, *, 3, 3, 3, 
    task_6:  *, *, *, 3, 3, 
    task_7:  *, *, 1, *, 1, 
    task_8:  *, *, *, 2, 2, 
    task_9:  *, *, 3, 3, 3, 
   task_10:  *, *, *, 3, 3, 
\end{verbatim}

{\bf  Permuted Mnist dataset.}

For Permuted Mnist all tasks tend to expand the first layer(linear1) while linear2 and linear3 are always shared. We use a shared task head.

\begin{verbatim}
columns: [linear1 linear2 linear3]
    task_0:  *, *, *,
    task_1:  *, 0, 0, 
    task_2:  *, 0, 0, 
    task_3:  *, 0, 0, 
    task_4:  *, 0, 0, 
    task_5:  *, 1, 1, 
    task_6:  *, 0, 0, 
    task_7:  *, 2, 2, 
    task_8:  *, 0, 0, 
    task_9:  *, 6, 6, 
\end{verbatim}

{\bf  VDD dataset.}

For VDD, most tasks create their own first three {\it conv} layers, and share the rest of the layers. 
We have omitted the {\it ImageNet} task because all the new layers in the Joint Net are initialized with corresponding {\it ImageNet} weights.
Training for the {\it Omniglot} dataset did not finish, hence we omit the corresponding task-net.

\begin{verbatim}
columns: [conv1 conv2 conv3 conv4 conv5 linear1 linear2]
         c100:  *, *, *,    *,    *,    *,    *
         svhn:  *, *, *, c100, c100, c100, c100, 
       ucf101:  *, *, *, svhn, svhn, svhn, svhn, 
        gtsrb:  *, *, *, svhn, svhn, svhn, svhn, 
daimlerpedcls:  *, *, *, svhn, svhn, svhn, svhn, 
  vgg-flowers:  *, *, *, svhn, svhn, svhn, svhn, 
     aircraft:  *, *, *, svhn, svhn, svhn, svhn, 
          dtd:  *, *, *, svhn, svhn, svhn, svhn, 
\end{verbatim}

\comment{
{\bf MLTR dataset.}
The MLTR dataset, created by us, is a multi-lingual scene text recognition dataset, consisting of natural scenes with text in several languages spoken in the Indian subcontinent region. 
For our experiments, we consider the following languages: {\it hindi}, {\it arabic}, {\it bangla}, {\it kannada}, {\it telugu}; the dataset consists of 10k images for each language, along with the ground truth text.
We use the well-known Convolutional Recurrent Neural Network (CRNN) for recognizing text, with a series of CNN layers followed by two bi-directional LSTM layers.
We will make a public release of the MLTR dataset soon.

In the final shared network obtained by our algorithm, we note that all the tasks grow their task-specific first four {\it conv}(c) layers, while the majority of the next layers are shared.
We observe an interesting phenomenon with this network: new layers are added for the first batch normalization layer ({\it bn1}), even though the previous {\it conv4} layer and all following layers are shared among all tasks.
This shows that by using conflict scores based on the RSA algorithm, we can pick expansion layers discriminatively.

\begin{verbatim}
columns: [c1 c2 c3 c4 c5 bn1 c6 bn2 c7 BiLSTM1 BiLSTM2]
 hin:  *, *, *, *,   *, *,   *,   *,   *,   *,   *,
arab:  *, *, *, *, hin, *, hin, hin, hin, hin, hin,
 ban:  *, *, *, *, hin, *, hin, hin, hin, hin, hin, 
 kan:  *, *, *, *, hin, *, hin, hin, hin, hin, hin, 
 tel:  *, *, *, *, hin, *, hin, hin, hin, hin, hin, 
\end{verbatim}
}


\subsection{Comparison with Dynamic Expansion Models.}
\begin{figure}[h]
    \centering
    \includegraphics[width=7cm,height=4cm]{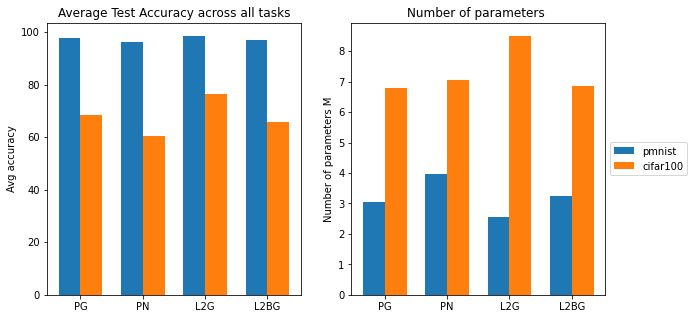}
    \caption{Comparison with other expansion methods on Permuted Mnist and Split-CIFAR100 tasks.}
    \label{fig:related}
\end{figure}

We compare our method with other recent expandable networks continual learning approaches PG\cite{pgn}, PN\cite{fernando2017pathnet}, L2G\cite{l2g}. We evaluate each compared approach by considering average test accuracy on all the tasks, model size and training time. Model size is measured via the number of model parameters after training all the tasks. We evaluate on Permuted Mnist and Split-CIFAR100 tasks, which are the only tasks common among these approaches.
Fig.~\ref{fig:related} shows the results.
We omit comparisons to regularization based approaches to continual learning~\cite{ewc,gem}, as they have been shown earlier to perform worse against the expansion based methods~\cite{l2g}.
Our approach (L2BG) has comparable average accuracy with previous approaches.
Compared with Learn-to-grow (L2G), L2BG uses more parameters to obtain similar performance.
L2G performs expensive optimization over all layer expansions, and hence is able to obtain more compact architectures.
In contrast, we employ a cheaper, approximate RSA-based conflict model to expand a subset of layers greedily, which leads to higher layer expansion. 
In future, we plan to explore other representation similarity techniques to compute improved conflict models~\cite{cca18} to help us compute smaller networks.
While L2BG does not improve performance on these tasks significantly, we argue that its advantages are orthogonal. 
The network architectures obtained by L2BG reflect similarity between different tasks and different layers thereof. Further, L2BG enables exploring the performance-size trade-offs (cf. Sec.~\ref{sec:bo}), which is not possible with previous expansion based approaches.

\subsection{Network Size-Performance Trade-offs}
\label{sec:bo}

\begin{figure}[ht]
    \centering
    \includegraphics[width=8cm]{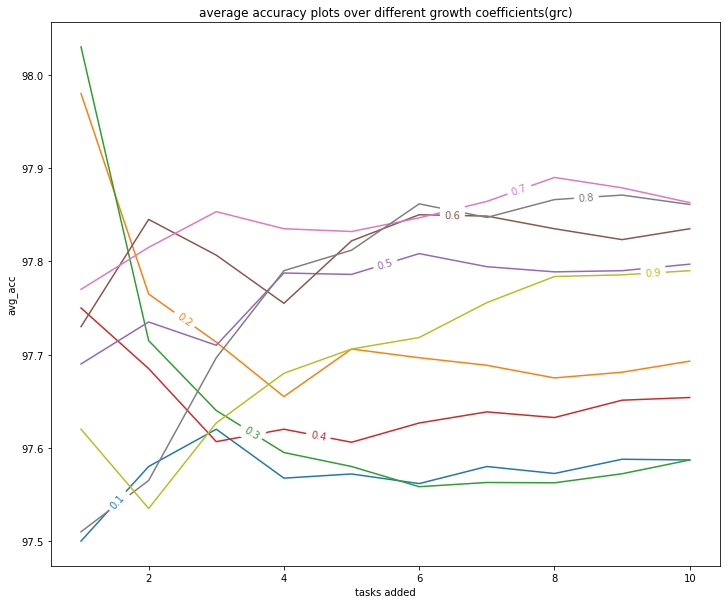}
    \caption{Average accuracy plots over different values of growth coefficient for Permuted Mnist }
    \label{fig:pmnist-acc}
\end{figure}

\begin{figure}[ht]
    \centering
    \includegraphics[width=8cm]{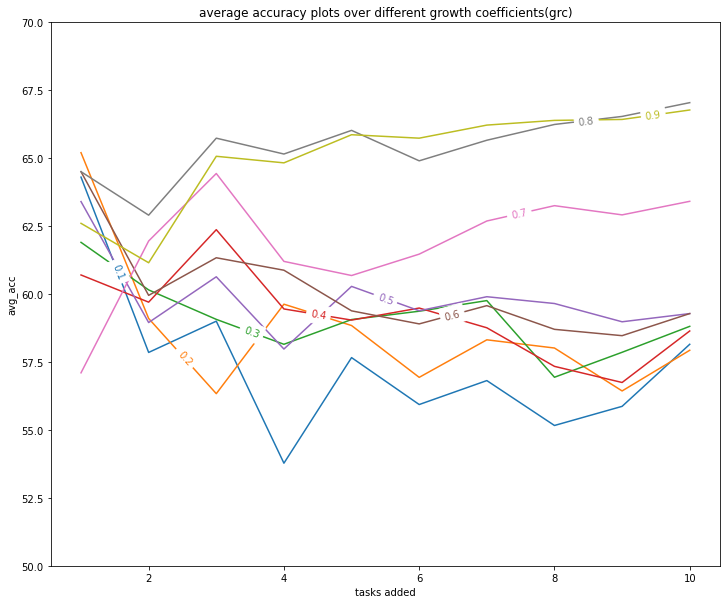}
    \caption{Average accuracy plots over different values of growth coefficient for Split-CIFAR100 }
    \label{fig:cifar-acc}
\end{figure}

\begin{figure}[ht]
    \centering
    \includegraphics[width=8cm]{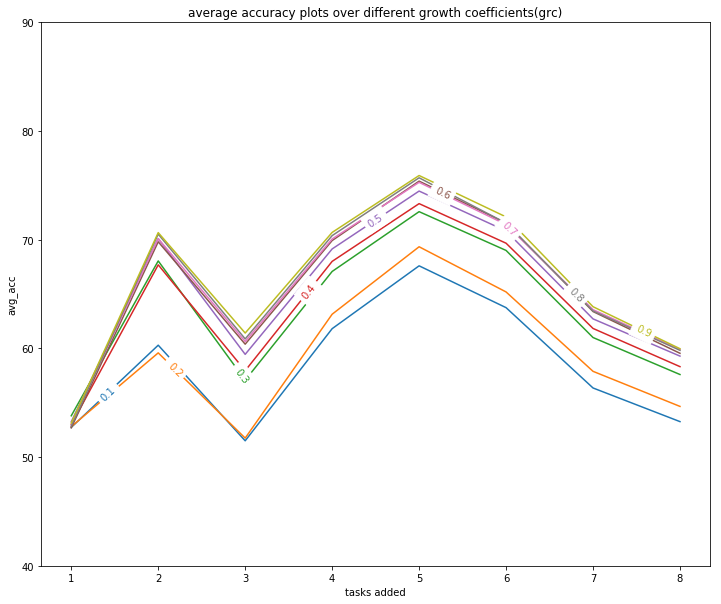}
    \caption{Average accuracy plots over different values of growth coefficient for VDD }
    \label{fig:vdd-acc}
\end{figure}
\comment{
\begin{figure}[ht]
    \centering
    \includegraphics[width=8cm]{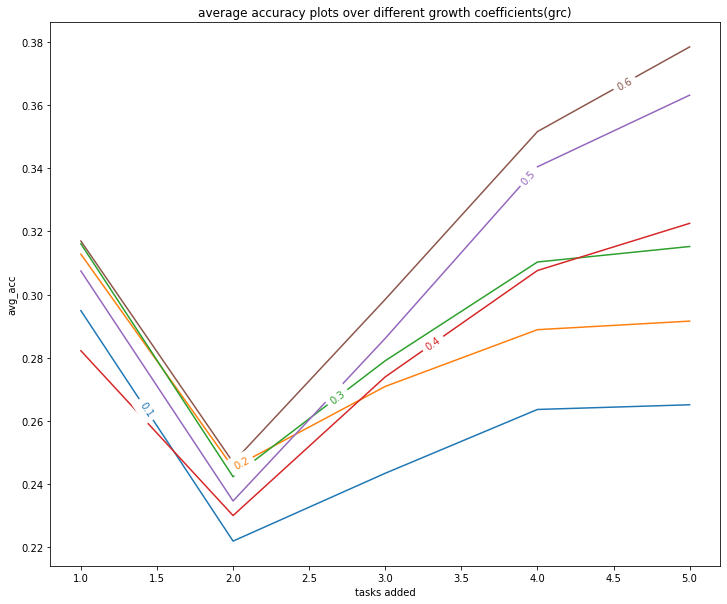}
    \caption{Average accuracy plots over different values of growth coefficient for MLTR }
    \label{fig:mltr-acc}
\end{figure}
}

In order to understand how grow coefficients impact the size and performance of the final architecture, we perform a grid search over grow coefficient values and evaluate the architectures obtained after arrival of each task.
For each coefficient value, L2BG computes task bindings and performs dynamic expansion: higher coefficient allows more expansion.
Fig.~\ref{fig:pmnist-acc},~\ref{fig:cifar-acc},~\ref{fig:vdd-acc}
(Split-CIFAR100, Permuted Mnist, VDD datasets, resp.) show how the average accuracy of the trained network changes as tasks arrive, for different growth coefficient values $\grt\in[0,1]$ with step size of $0.1$.
For each coefficient value, the final computed architecture corresponds to the right-most data point.

We observe that, in general, higher $\grt$ values lead to higher performance.
This leads to the size-performance tradeoff: higher $\grt$ networks use much more parameters (less sharing) to obtain high performance.
Interestingly, there are notable exceptions.
For example with Permuted Mnist dataset, we obtain the most performant network with $\grt = 0.7$; the network with $\grt = 0.9$, surprisingly, performs worse.
Similarly, for the VDD dataset, the final average accuracy for values in the range $\grt\in [0.5,0.9]$ are similar.
Thus, exploring the space of size vs performance is important to obtain a Pareto-optimal network: we may obtain a network with much fewer parameters, at the cost of a small drop in accuracy only.


{\bf Bayesian Optimization.} We also investigate the effectiveness of Bayesian Optimization (BO) in yielding global, Pareto optimal solutions.
Given the cost of running several growth sequences is high, we investigate if BO can yield useful network size-performance tradeoffs, even under low budget scenarios.
We use the Optuna framework~\cite{optuna19}, which uses the NSGA-II algorithm~\cite{nsga2} for multi-objective Bayesian Optimization.
To restrict the large optimization space of our benchmarks for a low budget BO setting, we assume a fixed task arrival order, keep a shared growth coefficient across tasks, and sample coefficient $\grt$ from a discrete uniform distribution $\mathcal{U}\{0,1\}$ with steps of $0.05$.
We perform 20 grow-sequence trials, each trial taking about 11 minutes on a single NVIDIA Tesla K80 GPU.

\begin{figure}[hbt]
    \centering
    \includegraphics[width=9cm]{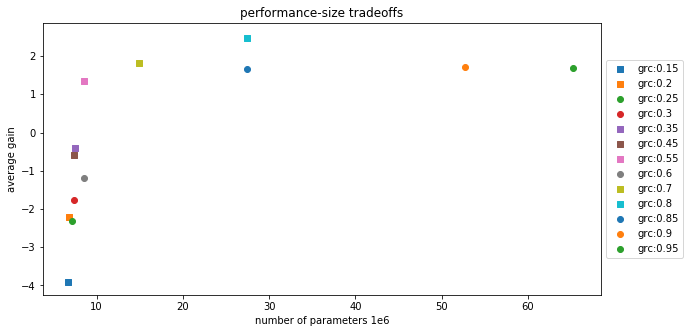}
    \caption{Trade-off between model size and performance for different values of grow coefficients. The boxes represent Pareto optimal solutions.}
    \label{fig:bo}
\end{figure}

Fig.~\ref{fig:bo} shows performance-size tradeoffs in the obtained solutions for the Split-CIFAR100 benchmark with varying $\grt$ (box-shaped points represent Pareto optimal solutions).
In general, large $\grt$ values ($\sim 0.95$) yield high average gains, but also yield a large parameter count.
Networks with fewer parameters are clustered on the left, with varying average (negative) gains.
We find a balanced solution at $\grt = 0.55$ with $< 10m$ parameters and average gain $1.34$.
Interestingly, we obtain the highest average accuracy solution at $\grt = 0.8$; the fully expanded network ($\grt \sim 0.95$) is inferior both in size (more than double) and performance.
Note that being able to visualize the Pareto optimal solutions enables us to make informed performance-size tradeoffs for downstream deployment scenarios.
No other existing expansion based method~\cite{den,pgn,l2g,rcl,apd20} is able to explore the space of optimal solutions globally.
In contrast to existing expansion based methods, which obtain a specific optimal architecture, we believe that the ability to visualize the space of optimal architectures is far more valuable.
Further, our novel parameterization of the network architectural space in terms of grow sequences enables global optimization even under low-budget scenarios.

\section{Conclusions}

We present a new framework for task-incremental learning, which learns to expand a joint network for multiple tasks dynamically, based on two new ideas: (a) exploit task similarity by binding an incoming task with the network of an existing, similar task, and (b) grow the layers relative to the similar task, by estimating a hierarchical conflict distribution between the layers of the similar and new tasks. The method enables multi-objective (performance, size) optimization across the space of shared, multi-task architectures.
There are several directions for future work: exploring techniques to scale BO to large multi-task benchmarks~\cite{taskonomy, mmnmt}, investigating other representation similarity techniques~\cite{cca18}, enabling expansion for non-symmetrical task networks, and adding layer adaptation to our algorithm.


\bibliographystyle{ACM-Reference-Format}
\bibliography{misc}


\end{document}